\documentclass{article}

\usepackage[preprint]{nips_2018}

\usepackage{xcolor}

\usepackage[utf8]{inputenc} 
\usepackage[T1]{fontenc}    
\usepackage{hyperref}       
\usepackage{url}            
\usepackage{booktabs}       
\usepackage{amsfonts}       
\usepackage{nicefrac}       
\usepackage{microtype}      
\usepackage{graphicx}
\usepackage{subcaption}
\usepackage{amssymb}
\usepackage{amsmath}
\usepackage{bm}

\def\eqref#1{equation~\ref{#1}}

\def\1{\bm{1}}

\def\eps{{\epsilon}}

\newif\ifMIT
\MITfalse
\ifMIT

\def\gamma{{\boldmath{\vgamma}}}

\else

\def\vgamma{{\bm{\gamma}}}

\fi

\ifMIT

\else

\fi

\DeclareMathAlphabet{\mathsfit}{\encodingdefault}{\sfdefault}{m}{sl}
\SetMathAlphabet{\mathsfit}{bold}{\encodingdefault}{\sfdefault}{bx}{n}

\ifMIT
\else

\fi

\hypersetup{
    colorlinks=true,
    linkcolor=blue,
    filecolor=magenta,      
    urlcolor=cyan,
    citecolor=blue,
}
\title{Unrestricted Adversarial Examples}

\author{Tom B. Brown\thanks{Google Brain. \space Correspondence to:  Tom B. Brown <tombrown@google.com>.}\And
Nicholas Carlini\footnotemark[1]\And
Chiyuan Zhang\footnotemark[1]\And
Catherine Olsson\footnotemark[1]\And
Paul Christiano\thanks{OpenAI.}\And
Ian Goodfellow\footnotemark[1]
}

\begin{document}

\maketitle

\begin{abstract}

We introduce a two-player contest for evaluating the safety and robustness
of machine learning systems, with a large prize pool.
Unlike most prior work in ML robustness, which studies norm-constrained adversaries,
we shift our focus to \emph{unconstrained} adversaries.
Defenders submit machine learning models, and try to achieve
high accuracy and coverage on non-adversarial data while making no \emph{confident mistakes} on
adversarial inputs.
Attackers try to subvert defenses by finding 
arbitrary unambiguous
inputs where the model assigns an incorrect label with high confidence.
We propose a simple unambiguous dataset ("bird-or-bicycle'')
to use as part of this contest.
We hope this contest will help to more comprehensively evaluate the
worst-case adversarial risk of machine learning models.
\end{abstract}

\section{Introduction}
\label{sec-intro}


Deep learning systems are superhuman at image classification
\citep{He16}, translation \citep{Wu2016} and game playing
\citep{Mnih13}.
However, good \emph{expected performance} can be
misleading: deep learning systems still make surprising and
egregious errors in the \emph{worst case} \citep{Szegedy2013}.
For example, image classifiers have been found to be
vulnerable to a range of attacks:
\begin{itemize}
    \item using optimization to make small pixel-wise modifications to clean
      inputs \citep{Szegedy2013};
    \item applying ordinary spatial transformations to the input (such
      as translations and rotations on images)
      \citep{Engstrom2017};
    \item and performing simple guess-and-check (trying a variety of inputs that seem
      likely to be misclassified) \citep{Gilmer2018}.
\end{itemize}

Even more troubling,
these systems often report \emph{high confidence} on
blatant errors \citep{nguyen2015deep}.
As a result, it is not currently
possible to protect against these problems simply by trusting the
system only when it is highly certain in its prediction.

In safety-critical situations, confident errors could be
catastrophic (e.g., the vision system behind a self-driving car
confidently believing there is no person in front of it when there is).
Even in situations where peoples' lives are not under
immediate risk, it is still be important for systems to know when
they are uncertain.\footnote{Indeed, there is a whole sub-field of
machine learning for estimating confidence in predictions. 
For an introduction, see \citep{Guo2017}. }
For instance, a home assistant might
think that it has heard its owner ask to text a sensitive message to some
unintended recipient.
If the system is not confident that it heard the recipient
correctly, it could ask for confirmation, but if it is incorrectly
confident then it could send a potentially-sensitive message to
someone who it was not intended to receive it (\citep{NYTimes2018}).
 
We believe it is important to design machine learning systems that 
\emph{never make confident mistakes}. \footnote{Note that it may only be possible to achieve this on some limited
set of tasks which are \emph{unambiguous}.}
Such a guarantee
would help protect not only against deliberate attacks by adversaries,
but also against unintentional accidents.

Is the goal of \emph{``no confident mistakes''} even possible? Two
aspects seem particularly challenging.
First, current models
consistently make confident mistakes even when the model input has
been only subtly changed.
Second, even if a model is able to resist a
certain attack, there is no guarantee that it will be robust to a
different attack.
For example, defending against an attacker who can make small modifications
to all pixels in an image does not protect against an adversary who
can make small rotations to an image
\citep{Engstrom2017}.
 
One might therefore conclude that no progress can be made towards the
goal of robustness to arbitrary inputs, especially given the limited progress
in the field of adversarial examples over the last several years.
We disagree.
In this paper, we
present two tools aimed at making this research direction tractable:


\begin{figure}
  \centering
  \includegraphics[width=0.7\linewidth]{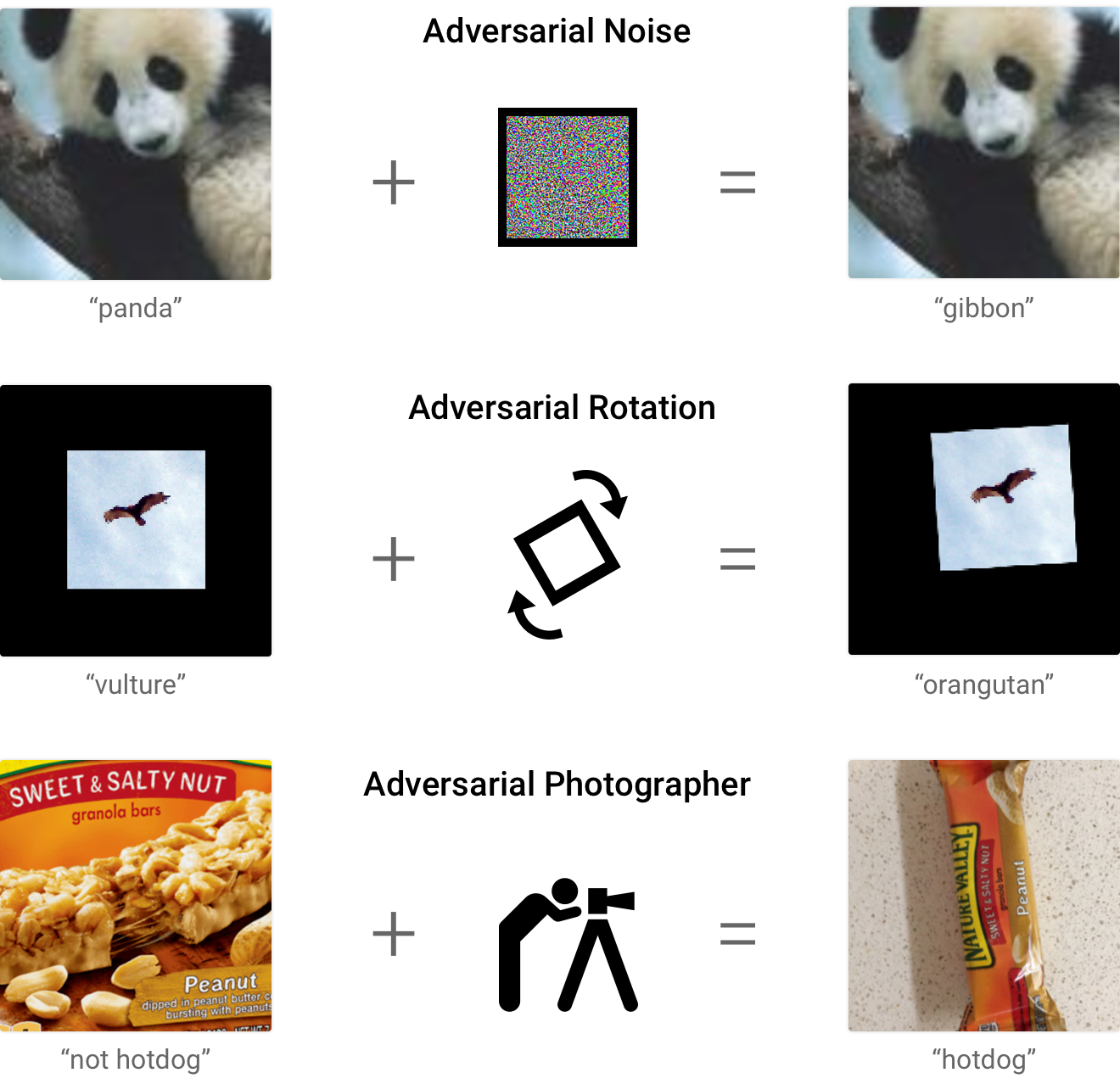}
  \caption{\textbf{Unrestricted adversarial examples} Adversarial
    examples can be generated in various ways, including small
    modifications to the input pixels, spatial transformations, or
    simple guess-and-check to find misclassified inputs.  
  }
  \label{fig-unrestricted-attacks}
\end{figure}

\subsection{Contest Overview}
We construct a two-class unambiguous dataset where all images are
either an unambiguous image of a bird or an unambiguous image of a bicycle.
We describe this dataset in Section~\ref{sec-task}.

Defenders build models that take an image as input, and must return either
\textbf{bird}, \textbf{bicycle} or \textbf{abstain}.\footnote{For an introduction to abstaining classifiers
and the  \emph{selective classification paradigm}, see \citet{Geifman2017}}
For inputs upon which the model does not abstain, the model must never make a mistake. 

Every few weeks, we ``stake'' a set of defenses that we believe 
have a chance of succeeding, and award a small
attacker prize to the first team who can break the defense.
To break a defense, an attacker must submit an unambiguous image of either
a bird or bicycle that the model labels incorrectly (as decided by an ensemble of human judges). 
Models are allowed to abstain (by returning low-confidence predictions)
on any adversarial input\footnote{
Models are prevented from always abstaining through the abstaining mechanism 
described in Section~\ref{sec-abstaining}}, but \emph{even one} confident error will result in a broken defense.

The first defense to remain un-broken for a $90$-day period will win a large
defender prize.

We will run a one-sided warm-up to the contest to test our infrastructure
and to ensure that there are defenses that can soundly defeat our fixed attacks. This warm-up is discussed in
Section~\ref{sec-warmup}.



\section{The Unambiguous Bird or Bicycle Task}
\label{sec-task}

We propose a two-way classification task for the purposes of
this challenge: ``Is this an unambiguous picture of a bird, a bicycle,
or is it (ambiguous / not obvious)?''.
We call this the
"bird-or-bicycle" task.

\begin{figure}
  \centering
  \includegraphics[width=1.0\linewidth]{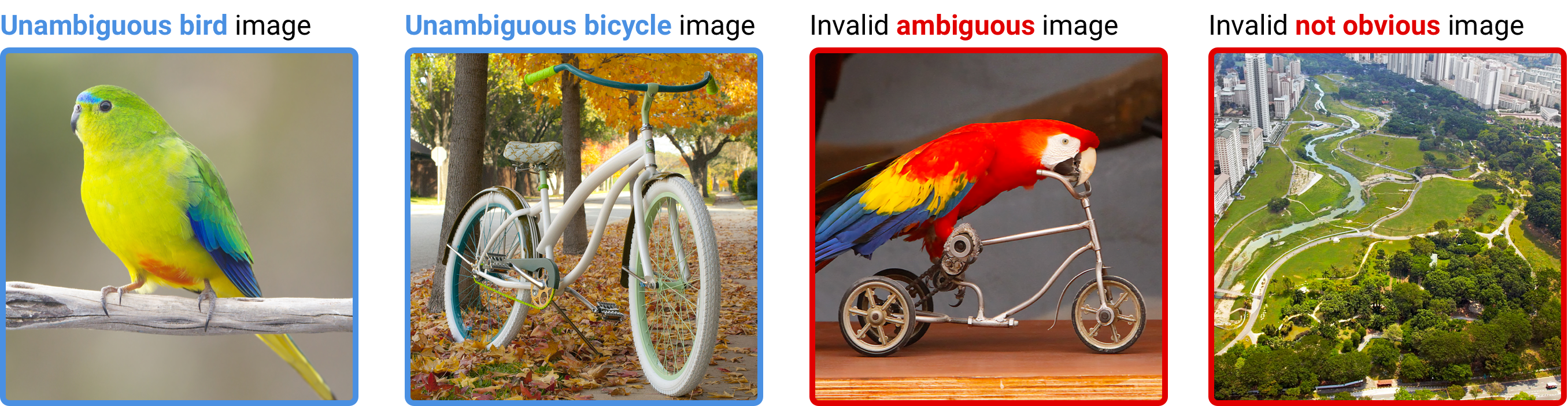}
  \caption{\textbf{Examples of candidate images for the
      bird-or-bicycle task.} The left two images are valid images for
    the bird-or-bicycle dataset. The right two images are invalid and
    not included in the dataset.
  }
  \label{fig-example-images}
\end{figure}

We selected this task because (a) given an arbitrary image, 
humans can robustly determine the correct label, (b) modern machine learning
techniques can also achieve perfect accuracy on an IID test set, but (c) the task is not so
simplistic that
it is possible to solve without machine learning
(see Appendix \ref{appendix-task-desiderata} for more detail).

We collect our images from the OpenImages
\citep{openimages2017} dataset; however, we intend the bird-or-bicycle task to
be well-defined on arbitrary images.
To obtain the ground truth label
for an image, we ask multiple human \emph{taskers} to label the
image as either a \textbf{bird}, a \textbf{bicycle}, or as being
 \textbf{ambiguous / not obvious}.
Any images where taskers did not unanimously agree the image is obviously either a bird or
bicycle are excluded from the dataset.
We provide some examples in Fig.~\ref{fig-example-images} and
full instructions provided to the taskers are shown in Appendix
\ref{appendix-tasker-instructions}.

%
%

\section{Phase One of the Challenge: The Warm-up}
\label{sec-warmup}
Before running the full unrestricted adversarial examples contest,
we would like to ensure that it is possible to solve this simple,
unambiguous task for a set of state-of-the-art \emph{fixed} attacks.
We have selected three attacks for this warm-up, including two black-box
pixel-wise optimization attacks \citep{Uesato2018,Brendel2017} and a
spatial attack \citep{Engstrom2017}.

Because these are fixed and restricted attacks, we expect that there will be
many ``uninteresting defenses'' that are tailored to perform well against
these specific fixed attacks, but will not generalize to other
attacks.\footnote{We give an example of one such uninteresting defense in
Appendix \ref{appendix-uninteresting-defenses}.}
Further, we expect that many defenses will unintentionally over-fit against
these specific attacks and not generalize to other attacks.
This further serves to motivate the full unrestricted two-player challenge.

After the research community can conclusively beat these fixed attacks, 
we will begin the two-sided contest phase.
Interested
parties can \href{https://github.com/google/unrestricted-adversarial-examples}{learn
  more about the warm-up on github}.

\section{Phase Two of the Challenge: The Contest}
\label{sec-contest}

\subsection{Differences from prior adversarial examples contests}
We propose a two-sided adversarial examples contest.
Unlike prior contests (e.g., the 2017 NIPS Adversarial Examples 
Contest \citep{Kurakin2018}),
our contest (a) is ongoing with no fixed end-date, (b) is fully
white-box, and (c) allows completely unrestricted attacks.

\paragraph{Our contest is ongoing, rather than fixed length.} 
We release new defenses that have been proposed every few weeks, 
and release attacks at the end of the week that they are submitted.
The rapid release cycle allows many rounds of back-and-forth between
attackers and defenders, which more closely resembles the dynamics of
systems deployed in the real world, and 
also allows contestants to individually verify that the submitted
adversarial images are in fact unambiguous.

When assigning the final defense prize, the organizers
will re-evaluate all the attack attempts to verify that they have all
failed.

Only when a defense has remained undefeated for several months will
we award the final defense prize.

\paragraph{Our contest is fully white-box, and does not rely on
oblivious adversaries.} 
Because the goal of this contest is to find models that make
\emph{no} confident mistakes (and not just for which it is hard for an
attacker to find mistakes) we require that all defenses are 
open-source and
available to the attackers.
Thus, attackers have the option to craft adversarial examples that target
specific models.
This provides a realistic threat
model for evaluating defenses, as recommended by \cite{Athalye2018}.

\paragraph{Our contest is unrestricted, and not constrained by any norm-ball.} 

As a consequence of not restricting attacks to a neighborhood around a
pre-labeled test point, each image must be evaluated by human judges
to obtain a true class.
An attack is successful if all the
judges agree that it is unambiguously class $A$ and the model
mistakenly classifies it as class $B$ with high confidence (instead of class $A$ or
abstaining).

\subsection{Abstaining mechanism for the contest}
\label{sec-abstaining}

In the contest, defenders attempt to create a model that never 
makes a confident mistake. 
Although the task is binary (A \emph{vs.} B), models are allowed to output three labels:
``Class A'', ``Class B'', or ``Abstain''.
We define a \emph{confident mistake} 
as the model assigning one class label, when
a unanimous ensemble of human taskers give the other class label.
Abstaining on all adversarial inputs is acceptable, and we allow defenses
to choose when to abstain using any mechanism they desire.

To prevent models from abstaining on \emph{all} inputs, models must
reach 80\% accuracy on a private \emph{eligibility dataset} of clean
bird-or-bicycle images. We reserve the right to perform additional tests to ensure defenders
do not over-fit against our private held-out data \citep{NYTimes2015}.

\subsection{The contest mechanics are open-sourced} 
We have open-sourced our proposed contest mechanics and are soliciting
feedback on them before launching the contest.
Readers can find
\href{https://github.com/google/unrestricted-adversarial-examples/blob/master/contest_proposal.md}{the
  most up-to-date contest proposal on GitHub}, and we also include a
snapshot of the contest proposal in Appendices
\ref{appendix-contest-proposal}, \ref{appendix-contest-overview} \&
\ref{appendix-contest-prizes}.

\section{Conclusion}
\label{sec-conclusion}

We have presented the unrestricted adversarial examples challenge,
consisting of a high-dimensional image classification task, and an
ongoing contest to evaluate defenses.
We hope that this challenge will encourage more researchers to 
investigate the difficult and conspicuously open problem of building
robust classifiers.

We invite interested parties to
test their defenses in the warm-up, and to give feedback on the
\href{https://github.com/google/unrestricted-adversarial-examples}{detailed
  contest mechanics in our open-source repository}.

\subsubsection*{Acknowledgments}
We thank the Hyderabad team for helping to create the bird-or-bicycle
dataset. We thank Alexey Kurakin for help in assembling early versions
of the dataset. We thank Justin Gilmer, Vincent Vanhouke, Jonathan
Uesato, Robert Obryk, Balaji Lakshminarayanan and Ananya Kumar for
useful conversations and feedback on early drafts of this paper.


\clearpage
\clearpage

\bibliography{paper}
\bibliographystyle{plainnat}

\clearpage
\clearpage

\appendix

\section{Contest proposal}
\label{appendix-contest-proposal}

The following appendix is a snapshot of our
\href{https://github.com/google/unrestricted-adversarial-examples/blob/master/contest_proposal.md}{living
contest proposal document on Github}.
We expect some specifics below to change as we refine the exact mechanics
of the contest, and we are actively seeking feedback on all details.
We are especially interested in any corner-cases that might allow for
attackers or defenders to ``cheat'' and win prizes without actually
following the spirit of the contest outlined previously.

\section{Contest overview}
\label{appendix-contest-overview}

All known machine learning systems make confident and blatant errors when 
in the presence of an adversary.
We propose an ongoing, two-player contest for evaluating the safety and robustness
of machine learning systems, with a large prize pool.
Unlike most prior work in ML robustness, which studies norm-constrained adversaries,
we shift our focus to \emph{unconstrained} adversaries.

Defenders submit machine learning models, and try to achieve
high accuracy and coverage on non-adversarial data while making no \emph{confident mistakes} on
adversarial inputs.
Attackers try to subvert defenses by finding 
arbitrary unambiguous
inputs where the model assigns an incorrect label with high confidence.

\subsection{Unambiguous two-class bird-or-bicycle dataset}
This contest introduces a new image dataset. 
We ask models to answer the question
 ``Is this an unambiguous picture of a bird, a bicycle,
or is it (ambiguous / not obvious)?''.
We call this the
"bird-or-bicycle" task.

We collect our images from the OpenImages
\citep{openimages2017} dataset.
To obtain the ground truth label
for an image, we ask multiple human taskers to label the
image as either a \textbf{bird}, a \textbf{bicycle}, or as being
 \textbf{ambiguous / not obvious}.
Only when all taskers unanimously agree the image is obviously either a bird or
bicycle is the
image included in the dataset.
(Details of the process are provided in Appendix~\ref{appendix-tasker-instructions}.)

\subsection{Defenders}
Defenders build models that output one of three labels: (a) bird, (b) bicycle, and (c) abstain.
Note that in the contest, researchers can choose to design any mechanism for abstaining
that they desire (unlike in the warm-up where we require that the defending model returns
two logits).

The objective of this contest is for defenders to build models that will never make
\emph{confident mistakes}.
By this, we mean that given any arbitrary input
that is unambiguous either a bird or a bicycle, the model must either
classify it correctly, or must choose to abstain.
To prevent models from vacuously abstaining on every input, the model may only abstain
on $20\%$ of an eligibility set (described below), but may abstain on any other input. 


\subsection{Attackers}
Attackers are given complete access to defenses (including the inference source code, 
training source code, and the pre-trained models).
Given this access, attackers attempt to construct 
unambiguous adversarial inputs that cause the
defense to make a \emph{confident mistake}.
There are two requirements here: 
\begin{enumerate}
\item the input must be unambiguously either a
bird or a bicycle, as decided by an ensemble of taskers (see Appendix~\ref{appendix-tasker-instructions}), and
\item the model must assign the incorrect label (and not choose to abstain).
\end{enumerate}
If both (a) all the human taskers agree the image is either an unambiguously
either a bird or bicycle,
and (b) the classifier assigns it the other label, then the attacker
wins.

The adversary will pass these adversarial images to the contest by uploading them to a
service run by the organizers.

\section{Contest prize distribution scheme}
\label{appendix-contest-prizes}
To incentivize actors to participate in this contest as both attackers and
defenders, we will provide a prize pool.

\subsection{Defender prize}
Half of the prize pool will be allocated for defenders.

The first defense that goes unbroken after
having been \emph{staked} (defined below)
for 90 days will win half of the total prize pool 
(the exact amount will be decided in the future).
We may allocate smaller prizes for
defenses that remain unbroken for smaller numbers of days.

Any researchers associated with the contest can submit defenses
but are ineligible for prizes.

\subsection{Attacker prizes}

We will allocate the other half of the attack prize for the attackers.

Attackers can submit adversarial instances to staked defenses at any time.
At the end of each week (11:59:59 PM on Friday) we will collect all adversarial
examples submitted that week and publish them publicly.
We will then send the images to taskers to confirm they are 
valid (i.e., are unambiguously a picture of a bird or a bicycle).
If attacks are found to be valid, then the defenders are notified
and can appeal the image if they disagree with the taskers.
A small monetary prize will be awarded to any team that breaks a
previously-unbroken defense with an eligible input.
If a defense is broken by
multiple attacks from the same week, then the prize is split among the
teams that beat the defense.

Because we need human evaluation of attacks, we allow up to 10 attacks per
day from any team with an email that is associated with a published
arXiv paper.
Researchers interested in participating who have not submitted an arXiv
paper previously can email the review board.
Any researchers associated with the contest can
submit attacks but are ineligible for prizes.

\subsection{Eligible defenses}
We want to avoid having people submitting defenses that are not novel
or real contenders to win the contest.

New defense submissions must do the following in order to be eligible:
\begin{enumerate}
    \item Obtain perfect accuracy on $80\%$ of the eligibility dataset;
    \item Successfully defend against all previously-submitted adversarial examples;
    \item Run in a docker container and conform to the API maintaining a throughput
      of at least 1 image per minute per P100 GPU;
    \item Be written in readable TensorFlow, PyTorch, Caffe, or pure NumPy;
      obfuscated or exceptionally hard-to-reproduce defenses are
      ineligible (as judged by a review board).
\end{enumerate}

Anyone can submit a defense at any time, and if it is eligible, we
will publish it by the end of the following week. Newly published defenses will initially
be unstaked, with no prize (other than honor) for breaking them.

\subsection{Staking an eligible defense}
A \emph{staked defense} has a small prize for attackers associated with it, and is also 
eligible to win the large defender prize if it remains unbroken for 90 days.

Defenses can become staked through one of two ways:

\paragraph{1.} A review board stakes the top defenses each week.

Every few weeks, a review board chooses the best
defenses that have been submitted, and stakes them with a small monetary prize.
Due to financial constraints we do not expect to be able to stake all defenses.

The criteria the review board will use for selecting defenses to stake are as follows

\begin{itemize}
\item The defense is clearly reproducible and easy to run;
\item The review board believes that it has a non-zero chance of claiming the Defender prize
(i.e., the proposal has not been broken previously);
\end{itemize}

There is some flexibility here. For example, 
preference will be given to defenses that have been accepted to peer-reviewed venues 
and the review board may stake a high-profile defense that has been accepted at a top
conference even if it is hard to run.
Conversely, the review board expects to stake defenses that are easy to run 
before they have been accepted at any conference.)

All the attacks that target staked defenses are released together.
If several defenses that were staked in the same batch remain unbroken for the
time period specified necessary to win the defense prize, it will
be split among them.

\paragraph{2.} Any team may pay to stake their own defense themselves.
The amount paid will be equal to the monetary prize that covers the prize
that will be awarded to the attacker who breaks the defense.
If a team pays to have their defense staked, and the defense is never broken, they
will receive their initial payment back along with their portion of the defender prize.

\subsection{Review Board}

We will form a \emph{review board} of researchers who helped organize the contest.
In exceptional circumstances, the review board has the power to withhold prizes (e.g.,
in cases of cheating)
No member of the review board will be eligible to win any prize.
Members of the review board will recuse themselves when discussing submissions from
researchers they conflict with (e.g., co-authored a recent paper, conflicted
institutionally, advisor/advisee, etc).

\subsection{Appeals}
Defenders who find images ambiguous that are labeled by taskers as unambiguously one class
can be appealed to the review board.
The review board will either have additional taskers label the image, or make an executive
decision.
In exceptional cases, an attacker can also appeal image labeled as ambiguous; however,
we expect it will be easier in most cases to re-upload a similar adversarial image to be
re-processed by taskers.

%
%
%

\section{An upper bound on the fraction of $L_\infty$ bounded images}
\label{appendix-lp-estimation-for-lulz}

We estimate the fraction of all images that lie within the $L_\infty$
neighborhood of any datapoint in the ImageNet test set.

\paragraph{How many 229 by 229 pixel images are there?}
For the ILSVRC 2014 challenge, each image contains $229^2$ pixels, and
each pixel can take on 256 values.
Thus we can calculate the total
number of images that conform to this specification.

$$
N_\mathrm{images} = 256^{229^2}
$$

\paragraph{How many images are in the $L_\infty$ attack neighborhood of each test point?}

Lets assume the standard norm ball size of $\eps=16$ (images have a
maximum pixel value of 255). 
Each image contains $229^2$ pixels and
each pixel can take on at most 32 values (+/- 16 values for each
pixel).
We therefore can use $32^{229^2}$ as the upper-bound of the
number of images images in the neighborhood of each test point.
$$
\forall x: N_\mathrm{neighborhood} (x) \leq 32^{229^2}
$$

\paragraph{How many total images are in the  $L_\infty$ attack neighborhood of at least one test point?}

The ILSVRC 2014 challenge has 40152 test images. Let's assume that a
negligible fraction of the images have overlapping neighborhoods. 
This gives us the following
$$
N_\mathrm{L_\infty} \leq \sum_{x_i}N_\mathrm{neighborhood}(x_i) \leq 40152 * 32^{229^2}
 $$
 
\paragraph{What fraction of images are within the neighborhood of one test-set point on ImageNet?}
Putting together our estimates from the preceding sections, we get the
following upper bound on the fraction of $L_\infty$ bounded images.
$$
\frac{N_\mathrm{L_\infty}}{N_\mathrm{images}}
\leq \frac{40152 * 32^{229^2}}{ 256^{229^2}}
\approx \frac{1}{ 10^{4494}}
$$

Thus we find that the fraction of total images covered by the
$L_\infty$ threat model is vanishingly small.

\section{Uninteresting defenses that solve the warm-up }
\label{appendix-uninteresting-defenses}
Because the warm-up consists only of fixed attacks, we expect there will
be a wide range of defenses that easily defeat the specific fixed attacks
we selected, but don't actually solve the problem of adversarial examples
under the restircted perturbation budgets.
The motivates \emph{why} we the full challenge is a two-player situation.

In order to demonstrate that simple uninteresting defenses exist, we have
constructed one defense that completely stops both of the supplied epsilon-ball
attacks.
(Observe that it is not possible to cheat on the rotations and translations adversarial
attack because we generate examples through brute force.)

Our ``defense'' is as follows:
\begin{verbatim}
    def np_model(x):
      x = x + np.random.normal(0, .05, size=x.shape)
      logits = sess.run(logits, {x_input: x})
      return np.array(logits == np.max(logits, axis=1), dtype=np.float32)
\end{verbatim}

This defense has two pieces: (1) randomness, which breaks the decision-only
attack, and (2) gradient masking, which breaks SPSA.
Both of these facts are well-known failure modes for these types of attacks,
we describe why each is ``effective'' below.

Randomness prevents the decision-only attack from working because it
constantly shifts the location of the decision boundary very slightly.
For an attack that works by walking along the decision boundary,
if we move where exactly the boundary is from one run to the next, we
won't be able to move along it.

SPSA works by numerically estimating the gradient of the input image
with respect to the probabilities.
By clipping the values to be identically either $1$ or $0$, we prevent
any possible gradient signal from being revealed.

\section{Desiderata in selecting a task}
\label{appendix-task-desiderata}
In selecting a task, we wanted to avoid tasks where we expect
hand-crafted (i.e. non-machine-learning-based) approaches could solve
the tasks, because we're interested in studying the security of ML
systems in particular.

If we chose a task that a hand-crafted algorithm could easily solve
(such as reversing a string) and tried to add a constraint that
solutions were "required to be ML-based", then we would expect that
the best performing solutions would cluster at the edge of whatever
definition of "ML" we chose.

\section{Reviewing inputs with taskers}
To review inputs and classify them as either a bird, a bicycle, or
ambiguous, we utilize an ensemble of several \emph{taskers}: humans
that we have selected to review inputs manually.
We will ask them to confirm that a given image definitely contains one
class, definitely does not contain the other, the object is not
truncated or occluded, and is a real object and not a painting, sculpture,
or other depiction of the object.

We will ask at least three taskers for each submitted image.
If any tasker is not certain, the image is rejected.
We will provide the taskers with multiple examples and continuously
monitor their responses.

Taskers will given unique IDs and all images will be released along with
all tasker's ID and response.

The review board reserves the right to over-rule the taskers, but expects
to do so only in exceptional circumstances.
If this is done, the review board will publicly explain why the result
was over-ruled.

Before a defense can win the defense prize, the review board will examine
every submitted adversarial example and confirm that all images rejected 
by taskers are in fact invalid.

\subsection{Instructions given to taskers}
\label{appendix-tasker-instructions}

\emph{We provide the following instructions to taskers.}

Answer the following questions:

\paragraph{1.} Does this photo contain a bird, or a depiction of a bird (e.g., a
toy bird, a painting of a bird, a stuffed animal bird, a cartoon bird) anywhere
in the image?
\begin{itemize}
\item Definitely yes (and I am confident that no other tasker will guess "no")
\item I'm not sure, but my best guess is yes
\item I'm not sure, but my best guess is no
\item Definitely no (and I am confident that no other tasker will guess "yes")
\end{itemize}

\paragraph{2.} Does this photo contain a bicycle, or a depiction of a bicycle (e.g.,
a drawing of a bicycle, a model bicycle, a toy bicycle)
anywhere in the image?
\begin{itemize}
\item Definitely yes (and I am confident that no other tasker will guess "no")
\item I'm not sure, but my best guess is yes
\item I'm not sure, but my best guess is no
\item Definitely no (and I am confident that no other tasker will guess "yes")
\end{itemize}

\emph{If the tasker said that there definitely IS NOT one class, and
  there MAYBE IS the other class, then move on to the following
  additional questions.}

\paragraph{3.} For the largest single bird/bicycle in the image, 
label the pixels of the bird/bicycle. 
(If there are multiple that are the same size, then choose one at random)

\paragraph{4.} Please answer the following True/False statements about the labeled bird/bicycle
\begin{itemize}
\item This bird/bicycle is complete and not truncated. It does not go
  outside of the image at all
\item This bird/bicycle is not occluded by anything else. I can see all of the bird/bicycle
\item This is a picture of a \textbf{[real, live bird]} / \textbf{[real bicycle]}. It is not a painting, drawing,
  sculpture, toy, stuffed animal, or any other sort of depiction.
(It is okay if the object is a photorealistic rendering of a bird/bicycle.)
\end{itemize}

\emph{The image is determined to be unambiguous ONLY IF all taskers answered
  ``Definitely yes'' to one class, ``Definitely no'' to the other class, the
  largest object is at least half of the image, is not truncated, is not occluded,
  and is not a depiction of any sort.}

\end{document}